\documentclass[final]{cvpr}

\usepackage{times}
\usepackage{epsfig}
\usepackage{graphicx}
\usepackage{amsmath}
\usepackage{amssymb}

\usepackage{bbm}
\usepackage{multirow}
\usepackage{enumitem}

\usepackage[pagebackref=true,breaklinks=true,letterpaper=true,colorlinks,bookmarks=false]{hyperref}

\usepackage{color}

\def\Approach{FAPIS}

\def\block1{{$\sf block1$}}
\def\block2{{$\sf block2$}}
\def\block3{{$\sf block3$}}
\def\block4{{$\sf block4$}}

\def\cvprPaperID{3844} 
\def\confYear{CVPR 2021}

\begin{document}

\title{FAPIS: A Few-shot Anchor-free Part-based Instance Segmenter}
\author{Khoi Nguyen and Sinisa Todorovic\\
Oregon State University\\
Corvallis, OR 97330, USA\\
{\tt\small {\{nguyenkh,sinisa\}}@oregonstate.edu}
}

\maketitle

\thispagestyle{empty}

\begin{abstract}
This paper is about few-shot instance segmentation, where training and test image sets do not share the same object classes. 
We specify and evaluate a new few-shot anchor-free part-based instance segmenter (\Approach). Our key novelty is in explicit modeling of latent object parts shared across training object classes, which is expected to facilitate our few-shot learning on new classes in testing. We specify a new anchor-free object detector aimed at scoring and regressing locations of foreground bounding boxes, as well as estimating relative importance of latent parts within each box. Also, we specify  a new network for delineating and weighting latent parts for the final instance segmentation within every detected bounding box. Our evaluation on the benchmark COCO-$20^i$ dataset demonstrates that we significantly outperform the state of the art. 
\end{abstract}

\section{Introduction}\label{sec:intro}

This paper addresses the problem of few-shot instance segmentation. In training, we are given many pairs of support and query images showing instances of a  target object  class, and the goal is to produce a correct instance segmentation of  the query given access to the ground-truth instance segmentation masks of the supports. In testing, we are given only one or a very few support images with their ground-truth instance segmentation masks, and a query image in which we are supposed to segment all instances of the target class. Importantly, the training and test image sets do not share the same object classes. Few-shot instance segmentation is a basic vision problem. It appears in many applications where providing manual segmentations of all object instances is prohibitively expensive. The key challenge is how to conduct a reliable  training on small data.

\begin{figure}[h!]
    \centering
    \includegraphics[scale=0.2]{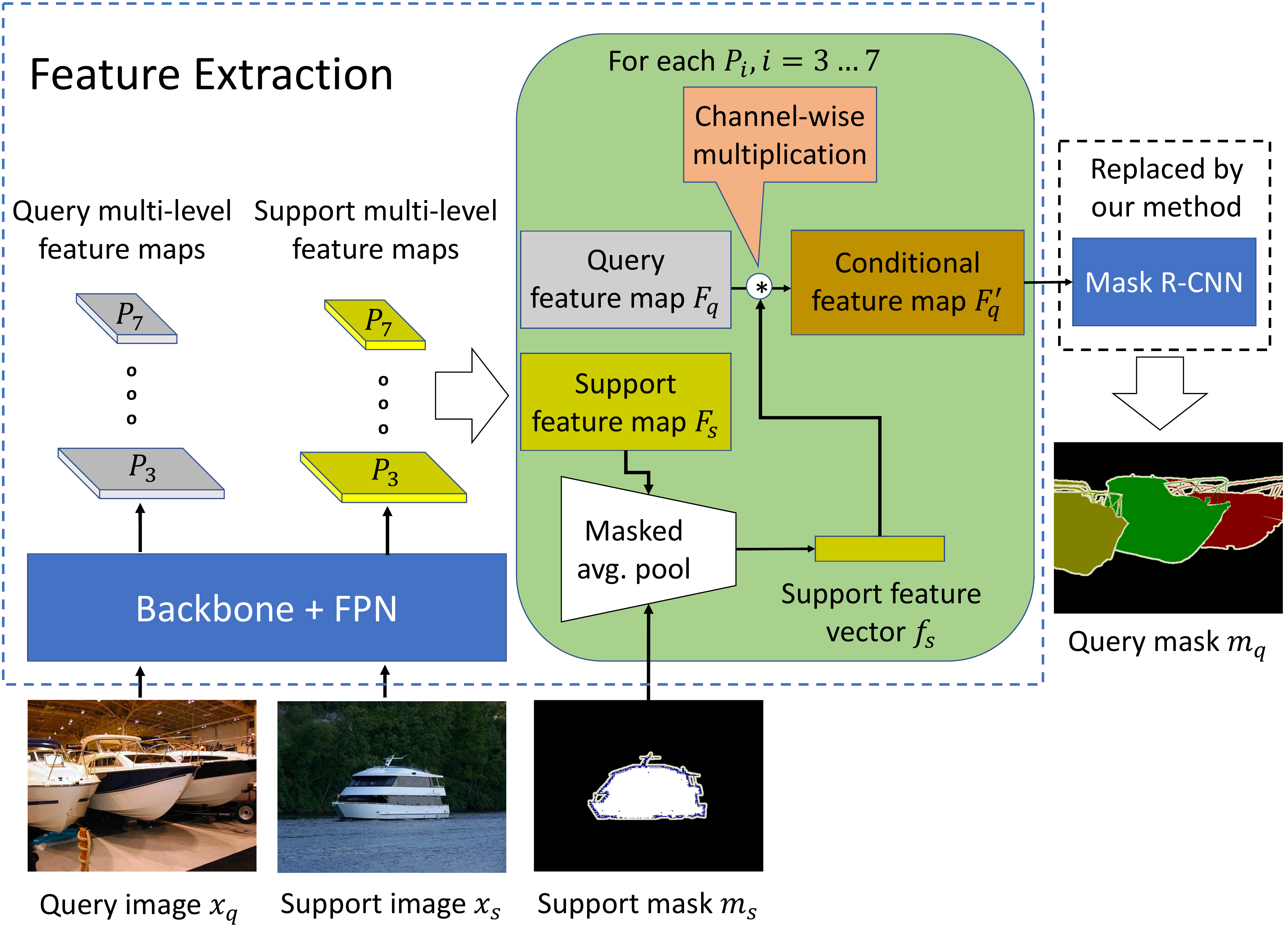}
    \caption{A common framework of prior work. The query and support image(s) are input to a backbone CNN with feature pyramid network (FPN) to extract the feature maps. The support's features modulate the query's features by channel-wise multiplication, resulting in the conditional query features, which are then input to Mask-RCNN for instance segmentation of the query. Our approach replaces Mask-RCNN with two new modules for anchor-free object detection, and part-based instance segmentation.}
    \label{fig:feature_extraction}
\end{figure}

Fig.~\ref{fig:feature_extraction} illustrates a common framework for few-shot instance segmentation that typically leverages Mask-RCNN \cite{he2017mask,  yanICCV19metarcnn, michaelis2018one}. 
First, the support and query images are input to a backbone CNN and feature pyramid network (FPN) \cite{lin2017feature} for computing the support's and query's feature maps. Second, for every feature map and every support's segmentation mask, the masked average pooling computes the support's feature vector. Third,
the support's feature vector is used to modulate the query's feature maps through a channel-wise multiplication, resulting in the conditional query feature maps. Finally, the conditional query features are forwarded to the remaining modules of Mask-RCNN to produce instance segmentation of the query.

\begin{figure*}[h!]
    \centering
    \includegraphics[scale=0.37]{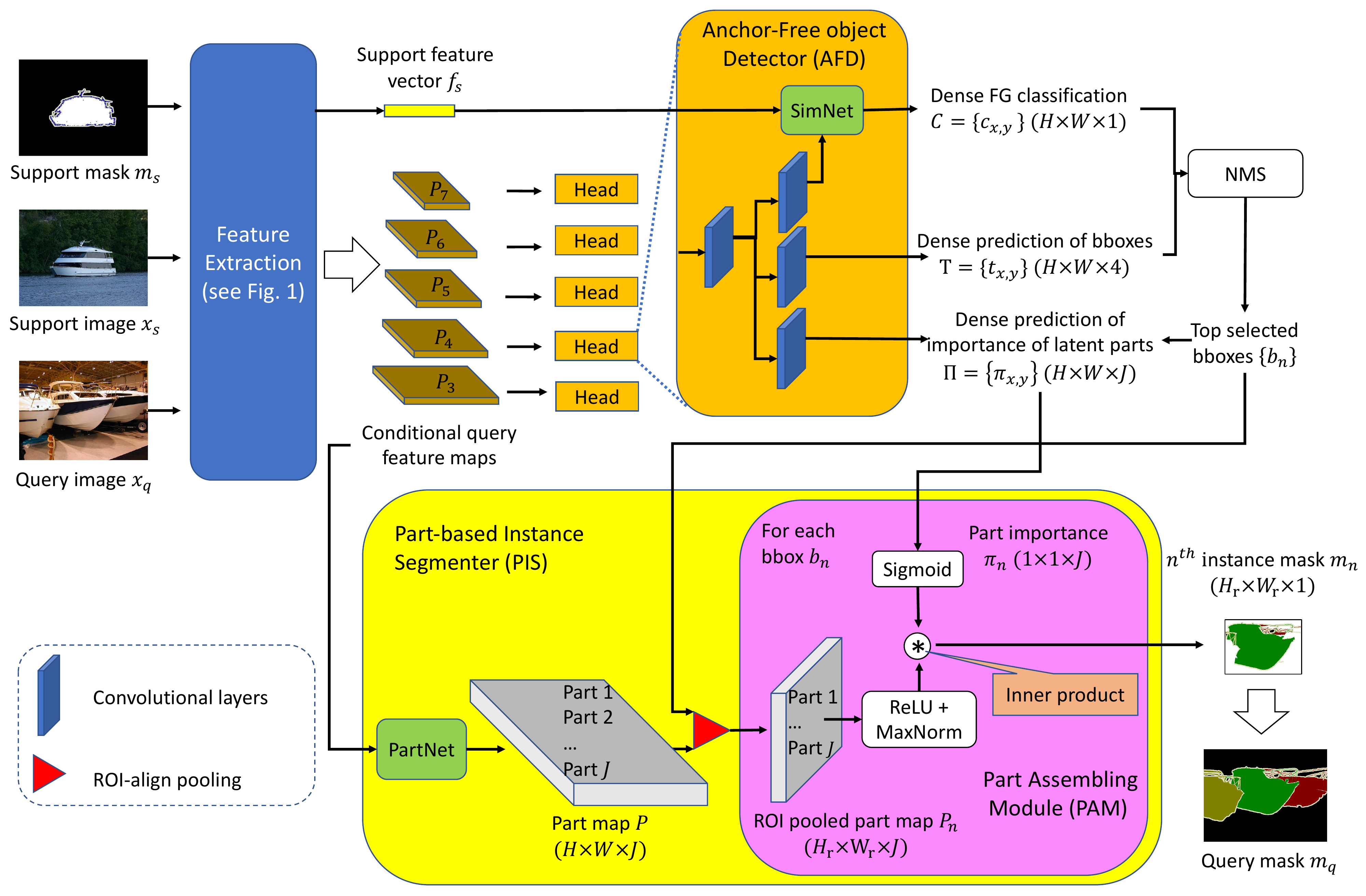}
    \caption{Our \Approach~uses the same feature maps as in Fig.~\ref{fig:feature_extraction}, and extends prior work with two new modules: anchor-free object detector (AFD) and  part-based instance segmentor (PIS). The AFD produces three types of dense predictions 
    for every location $(x,y)$ in the feature map: (a) Figure-ground (FG) classification score; 
    (b) Location of the closest bounding box to $(x,y)$; 
    (c) Relative importance of the latent parts for segmentation of the bounding box closest to $(x,y)$. 
    The PIS consists of the PartNet and Part Assembling Module (PAM). The PartNet predicts activation maps of the latent parts. After NMS selects the top scoring bounding boxes, for every box $n$,  the PAM fuses the part-activation maps according to the predicted part importance for the box $n$, resulting in the instance segmentation  $m_n$.}
    \label{fig:part_assembling_module}
\end{figure*}

This framework has limitations. First, Mask-RCNN is anchor-based, and hence might overfit to particular sizes and aspect ratios of training objects, which do not characterize new classes in testing. Second, the Mask-RCNN learns feature prototypes \cite{he2017mask} which are correlated with the feature map from the backbone in order to produce object segmentation. However, the prototypes  typically capture global outlines of objects seen in training \cite{he2017mask}, and hence may not be suitable for segmenting new object classes with entirely different shapes in testing.

To address these limitations, we propose \Approach~-- a new {\bf f}ew-shot {\bf a}nchor-free {\bf p}art-based {\bf i}nstance {\bf s}egmenter, illustrated in Fig.~\ref{fig:part_assembling_module}. In a given query, \Approach~first detects bounding boxes of the target object class defined by the support image and its segmentation mask, and then segments each bounding box by localizing a universal set of latent object parts shared across  all object classes seen in training. 

Our key novelty is in explicit modeling of latent object parts, which are smaller object components but may not  be meaningful (as there is no ground truth for parts). Unlike the prototypes of \cite{he2017mask}, our latent parts capture certain smaller components of objects estimated as important for segmentation. As parts may be shared by  distinct objects, including new object classes of testing, we expect that accounting for parts will lead to a more reliable few-shot learning than the aforementioned common framework. We are not aware of prior work that learns latent parts for few-shot instance segmentation.


We make two contributions. First, we specify a new {\em anchor-free object detector } (AFD) that does not use a set of candidate bounding boxes with pre-defined sizes and aspect ratios, called anchors, and, as shown in \cite{tian2019fcos}, in this way mitigates over-fitting to a particular choice of anchors. The AFD  (the orange box in Fig.~\ref{fig:part_assembling_module})  is  aimed at three tasks at every location of the query's  feature map: dense scoring and regressing locations of foreground bounding boxes, as well as dense estimation of a relative importance of the latent parts for segmentation.  While all of the latent parts are learned to be relevant for object segmentation, differences in sizes and shapes across instances will render some latent parts more important than some others for segmentation of each instance. Thus, the third head in the AFD estimates the part importance which varies across the predicted bounding boxes in the image, as desired. The AFD's output is passed to the standard non-maximum suppression (NMS) for selecting top scoring bounding boxes.

Second, we specify a new {\em Part-based instance segmenter} (PIS). The PIS (the yellow box in Fig.~\ref{fig:part_assembling_module}) is aimed at localizing and integrating latent object parts to produce the final instance segmentation within every NMS-selected  bounding box. The PIS consists of the PartNet and part assembling module (PAM). The PartNet predicts activation maps of the latent parts, called part maps, where large activations strongly indicate the presence of the corresponding part in the image. Importantly, high activations of a certain latent part at image location $(x,y)$ may not be important for segmentation of the object instance at $(x,y)$ (e.g., when several latent parts overlap but do not ``cover'' the full spatial extent of the instance). Therefore, for every NMS-selected  bounding box, these part maps are then integrated by the PAM so as to account for the predicted relative importance of the parts for that box. Finally, all instance segmentations form the output query segmentation mask.

Our evaluation on the COCO-$20^i$ dataset \cite{michaelis2018one} demonstrates that we significantly outperform the state of the art. 

In the following, Sec.~\ref{sec:related_work} reviews prior work; Sec.~\ref{sec:method} specifies our deep architecture; 
Sec.~\ref{sec:experiments} presents our implementation details and experimental results; and Sec.~\ref{sec:conclusions} describes our concluding remarks.

\section{Related Work}
\label{sec:related_work}



\textbf{Few-shot semantic segmentation} labels pixels in the query with  target classes, each defined by $K$ support examples  \cite{shaban2017one,rakelly2018conditional,rakelly2018fewshot,dong2018few,zhang2018sgone,simonyan2014deep, shelhamer2016fully,hu2019attention,siam2019adaptive,zhang2019canet,zhang2019pyramid,yang2020new,nguyen2019feature,wang2019panet}. 
Our problem is arguably more challenging than few-shot semantic segmentation, since we need to additionally distinguish object instances of the same class.


\textbf{Instance segmentation} localizes  instances of object classes seen in training, whereas we are supposed to segment instances of new classes. There are proposal-based and proposal-free approaches. The former  \cite{he2017mask,li2017fully,chen2018masklab,liu2018path,bolya2019yolact} first detects object bounding boxes, and then segments foreground within every box.  The latter  \cite{kirillov2017instancecut,kong2018recurrent,chandra2017dense,fathi2017semantic} typically starts from a semantic segmentation, and then leverages certain visual cues (e.g., object center, or Watershed energy) to cluster pixels of the same semantic class into instances. 

Our \Approach~and YOLACT \cite{bolya2019yolact} are similar in that they both predict activation maps of certain object features, and at each location in the image they fuse these activation maps by weighting their relative importance for segmentation. However, since YOLACT  is not aimed at the few-shot setting, there are a number of significant differences.
First, our \Approach~models latent object parts, whereas YOLACT models object prototypes representing global outlines of groups of similar object shapes. While the latter has been argued as suitable for instance segmentation, our experiments demonstrate that the prototypes learned on training examples poorly represent new, differently shaped object classes in testing for few-shot instance segmentation. As our latent parts may be components of the new object classes, they are likely to better represent new objects than the global shapes of prototypes. Second, YOLACT is standard anchor-based detector, whereas our AFD object detector is anchor-free producing dense predictions of bounding boxes. Our approach is more suitable for few-shot instance segmentation, as pre-defined anchors in training may not represent well the sizes and aspect ratios of new objects in testing.


\textbf{Few-shot instance segmentation} approaches usually adapt methods for instance segmentation (e.g., Mask-RCNN  \cite{he2017mask}) to the few-shot setting \cite{michaelis2018one,yanICCV19metarcnn,fan2020fgn} (see Fig.~\ref{fig:feature_extraction}). Our main differences are in replacing the Mask-RCNN with our AFD and PID for modeling latent object parts and segmenting every detected instance by localizing and assembling latent parts relevant for that instance.

\textbf{Anchor-free Object Detection} \cite{tian2019fcos,zhou2019objects,law2018cornernet} predicts bounding boxes for all pixels in the feature map. This is opposite to anchor-based approaches \cite{ren2015faster,lin2017focal} where a set of bounding boxes with pre-defined sizes and aspect ratios are classified as presence or absence. Our AFD follows the FCOS approach \cite{tian2019fcos}, where at each location $(x,y)$ distances to the top, bottom, left, right sides of the closest bounding box are predicted by regression. We extend FCOS with SimNet to more reliably score bounding boxes, and thus reduce the number of false positives.

{\bf Part-based segmentation.} While the literature abounds with approaches that use parts for image classification and object detection, there is relatively scant work on part-based segmentation. In \cite{trulls2014segmentation,girshick2014rich} the deformable part model \cite{DPM_PAMI09} is used for semantic segmentation, and in \cite{liu2017sgn} reasoning about ``left'', ``right'', ``bottom'' and ``top'' parts of the foreground object is shown to help instance segmentation. Recent work \cite{li2017holistic,gong2018instance} addresses the problem of human (one class) segmentation, not our multi-class segmentation problem. We are not aware of prior work that learns latent parts for few-shot instance segmentation.

\section{Our Approach}
\label{sec:method}

\subsection{Problem Statement}
In training, we are given many pairs of support and query images showing the same target class from the class set $\mathcal{C}_1$, along with their pixel-wise annotations of every instance of the target class. In testing, we are given $K$ support images, $\{x_s\}$, and their $K$ instance segmentation masks $\{m_s\}$ that define a target class sampled from the class set $\mathcal{C}_2$, where $\mathcal{C}_1 \cap \mathcal{C}_2 = \emptyset$. Given a test query, $x_q$, showing the same target class as the test supports, our goal is to segment all foreground instances in $x_q$, i.e., estimate the query instance segmentation mask $m_q$. This problem is called \textbf{1-way $K$-shot} instance segmentation. In this paper, we consider $K=1$ and $K=5$, i.e., test settings with a very few supports.
It is worth noting that we can easily extend the 1-way $K$-shot to $N$-way $K$-shot by running $N$ support classes on the same query, since every class can be independently detected. However, the $N$-way $K$-shot problem is beyond the scope of this paper.

\subsection{Multi-level Feature Extraction}
As shown in Fig.~\ref{fig:feature_extraction}, \Approach~first extracts multi-level feature maps at different resolutions from $x_s$ and $x_q$ using a backbone CNN with feature pyramid network (FPN), as in \cite{lin2017feature}. For each level  $i=3,4,...,7$, the support feature map $F_{s,i}$ is masked-average pooled within a down-sampled version of the support mask $m_s$ to obtain the support feature vector $f_{s,i}$. Then, for each level  $i$,  $f_{s,i}$ is used to modulate the corresponding query feature map $F_{q,i}$.  Specifically, $f_{s,i}$ and $F_{q,i}$ are multiplied channel-wise. This gives the conditional query feature map $F'_{q,i}$. The channel-wise multiplication increases (or decreases) activations  in the query's $F'_{q,i}$ when the corresponding support's activations are high (or low). In this way, the channel features that are relevant for the target class are augmented, and irrelevant features are suppressed to facilitate instance segmentation.

\subsection{The Anchor-free Object Detector}
For each conditional query feature map $F'_{q,i}$,  $i=3,4,...,7$,  the AFD scores and regresses locations of foreground bounding boxes. Below, for simplicity, we drop the notation $i$ for the feature level, noting that the same processing is done for each level $i$. The workflow of  the AFD is illustrated in the orange box in Fig.~\ref{fig:part_assembling_module}.

For every location $(x,y)$ in $F'_q$ with height and width $H\times W$,  the AFD predicts:
\begin{enumerate}[itemsep=-1pt,topsep=-1pt, partopsep=2pt]
\item Figure-ground (FG) classification scores $C=\{c_{x,y} \}\in [0,1]^{H\times W}$ using the SimNet and Classification Head;
\item Regressions $T=\{t_{x,y}\} \in \mathbbm{R}_+^{H\times W\times4}$ of top, bottom, left, right distances from $(x,y)$ to the closest box.
\item Relative importance of the $J$ latent parts for instance segmentation of the bounding box predicted at $(x,y)$, $\Pi=\{\pi_{x, y} \}\in \mathbbm{R}^{H\times W\times J}$.
\end{enumerate}

{\bf SimNet and Classification Head.} Prior work \cite{yanICCV19metarcnn,michaelis2018one},
uses the support feature vector to modulate the
query feature map by channel-wise multiplication, and thus detect target objects. However, in our experiments, we have observed that this method results in a high false-positive rate. To address this issue, we specify the SimNet which consists of a block of fully connected layers followed by a single convolutional layer. The first block takes as input $f_s$ and predicts weights of the top convolutional layer. This is suitable for addressing new classes in testing. Then, the top convolutional layer takes as input $F'_q$ and the resulting feature maps are passed to the FG head for predicting the FG classification scores $C$ at every location $(x,y)$. In this way, the SimNet learns how to more effectively modulate $F'_q$, extending the channel-wise multiplication in \cite{yanICCV19metarcnn,michaelis2018one}.

For training the FG head, we use the focal loss  \cite{lin2017focal}:
\begin{equation}
\label{eq:cls_loss_function}
    L_{c} = \frac{-1}{H \times W} \sum_{x,y}\alpha'_{x,y}(1 - c'_{x,y})^\gamma \log(c'_{x,y}),
\end{equation} 
where $c'_{x,y} = c^*_{x,y}c_{x,y} + (1-c^*_{x,y})(1-c_{x,y})$, $c^*_{x, y}\in\{0,1\}$ is the ground-truth classification score at $(x,y)$; $\alpha'_{x,y} = c^*_{x,y}\alpha + (1-c^*_{x,y})(1-\alpha)$, $\alpha \in [0, 1]$ is a  balance factor of classes, and $\gamma \in [0, 5]$ is a focusing parameter to smoothly adjust the rate at which easy examples are down-weighted. 

{\bf Bounding Box Regression.}
For training the box regression head, we use the general intersection-over-union (GIoU) loss \cite{rezatofighi2019generalized}: 
\begin{equation}
\label{eq:reg_loss_function}
    L_{b} = \frac{1}{N_{pos}} \sum_{x,y} {\mathbbm{1}(c^*_{x, y} = 1)} \;\cdot\; \text{GIoU} (t_{x,y}, t^*_{x,y}),
\end{equation} 
where $t^*_{x,y}\in\mathbbm{R}_+^4$ is a vector of the  ground-truth top, bottom, left, right box distances from $(x, y)$;  $\mathbbm{1}(\cdot)$ is the indicator function having value 1 if $(x,y)$ belongs to the ground-truth box and 0 otherwise; and $N_{pos} = \sum_{x,y} {\mathbbm{1}(c^*_{x, y} =1)}$ is the number of locations in the feature map that belong to the foreground.

{\bf Part-Importance Prediction Head.}
For every location $(x,y)$ in $F'_q$, this head predicts the relative importance of the latent parts $\Pi=\{\pi_{x, y} \}\in \mathbbm{R}^{H\times W\times J}$. This part-importance prediction seeks to capture the varying dependencies among the latent parts when jointly used to estimate a segmentation mask of the bounding box predicted for every $(x,y)$. Note that we are not in a position to specify an explicit loss for training this head, because the parts are latent, and not annotated in our training set. Importantly, we do get to train this head by backpropagating the instance segmentation loss, as described in Sec.~\ref{sec:PIS}.

The AFD's predictions are forwarded to the standard NMS to select the top scoring $n=1,\dots,N$ bounding boxes at all feature-pyramid levels. 

\subsection{The Part-based Instance Segmenter}\label{sec:PIS}

The PIS is illustrated in the yellow box in Fig.~\ref{fig:part_assembling_module}. The PIS consists of the PartNet and the PAM.

\textbf{The PartNet} takes as input the largest conditional query feature map -- specifically, $F'_{q,3}$  -- and predicts  activation maps of the $J$ latent parts, or part maps for short, $P \in \mathbbm{R}^{H \times W \times J}$. $J$ is a hyper-parameter that we experimentally set to an optimal value. High activations in $P$ for a particular latent part at location  $(x,y)$ strongly indicate that part's       presence at $(x,y)$. For every bounding box $n=1,\dots,N$ selected by the NMS, we conduct region-of-interest (ROI) aligned pooling of $P$ to derive the $J$ ROI-aligned pooled part maps $P_n\in \mathbbm{R}^{H_{\text{r}}\times W_{\text{r}}\times J}$, where $H_{\text{r}}$ and $W_{\text{r}}$ are the reference height and width ensuring that the pooled features of the $N$ bounding boxes have the same reference dimension.  The PartNet is learned by backpropagation of the instance segmentation loss through the PAM module, since the latent parts are not annotated and we cannot define an explicit loss for training the PartNet.

A strongly indicated presence of a certain latent part at location $(x,y)$ in $P_n$ may not be important for segmentation of the object instance in the bounding box $n$ (e.g., when this information is redundant as many other latent parts may also have a high activation at $(x,y)$ in $P_n$). For each location, we obtain a triplet of classification scores, bounding box and part importance $\pi_n$. Therefore, for every bounding box  $n=1,\dots,N$, the following PAM module takes as input both the pooled part maps $P_n \in \mathbbm{R}^{H_{\text{r}}\times W_{\text{r}} \times J}$ and the estimated part importance $\pi_n \in \mathbbm{R}^{J}$ for  segmentation of instance $n$.

\textbf{The PAM} computes the segmentation mask of every bounding box $n=1,\dots,N$, $m_n\in[0,1]^{H_{\text{r}}\times W_{\text{r}}}$, as%
\begin{equation}
    m_n =  P_n^+ \otimes \sigma (\pi_n),\quad
    P_n^+ =\text{MaxNorm}(\text{ReLU} (P_n)),
    \label{eq:PAM_function}
\end{equation} 
where $\text{MaxNorm} (A) {=} \frac{A}{\max_{x,y} A_{x,y}}$ for an activation map $A$,
 $\otimes$ denotes a inner product, and $\sigma$ is the sigmoid  function. 
 Note that  $\otimes$ in \eqref{eq:PAM_function} requires the tensor $P_n^+$  be rectified to a matrix of size $(H_{\text{r}} W_{\text{r}})\times J$. Thus,  the operator  $\otimes$ serves to fuse the part  maps by their respective importance for instance segmentation.

The  MaxNorm-ReLU  composite function and the sigmoid in \eqref{eq:PAM_function} make the part maps and their importance non-negative with values in $[0,1]$. This design choice is inspired by the well-known {\em non-negative matrix factorization} (NMF) \cite{lee1999learning}, so that if the instance segmentation matrix $m_n$ were available, the expression in \eqref{eq:PAM_function} would amount to the NMF of $m_n$ into the non-negative $P_n^+$ and $\sigma (\pi_n)$. This design choice is conveniently used to  regularize learning of the PartNet and the AFD's head for predicting the  part importance. Specifically, in learning, we maximize similarity between the predicted $P_n^+$ and the NMF's bases computed for the  ground-truth instance segmentation masks $\{m_n^*:n=1,\dots, D\}$, where $D$ is the total number of instances in the training set. This NMF-based regularization in learning enforces that our $P_n^+$ is {\em sparse}, justifying our interpretation that $P_n^+$ represents the smaller latent parts of object instances. 



{\bf Segmentation loss and NMF-based regularization.}
The prediction $m_n$, given by \eqref{eq:PAM_function}, incurs the following segmentation loss $L_{s}$ with respect to the ground-truth instance segmentation mask $m_n^*$: %
\begin{equation}
\begin{array}{rcl}
    L_{s} &=& \displaystyle \frac{1}{N} \sum_{n=1}^N l_{dice}(m_n, m^*_n),
    \\
    l_{dice}(A,B) &=& 1 - \frac{2 \sum_{x, y}A_{x, y} \cdot B_{x, y}}{\sum_{x, y} A_{x, y}^{2}+\sum_{x, y} B_{x, y}^{2}},
\end{array} 
    \label{eq:mask_loss}
\end{equation}
where $l_{dice}$ is the  dice loss \cite{milletari2016v} between maps $A$ and $B$. 

\begin{figure}[h!]
    \centering
    \includegraphics[scale=0.3]{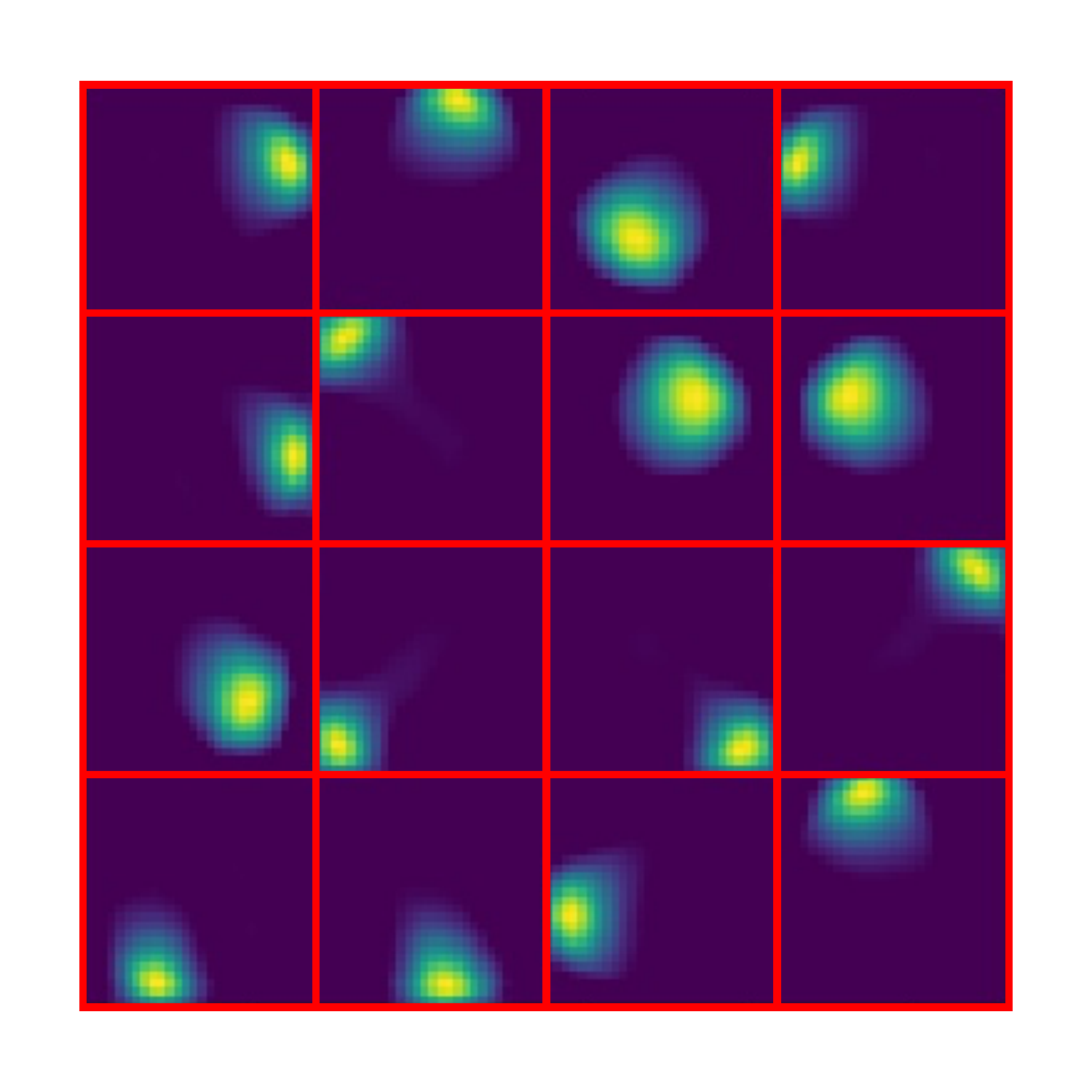}
    \caption{Visualization of 16 NMF parts estimated from the size-normalized ground-truth instance masks in COCO-$20^i$ \cite{michaelis2018one}.}
    \label{fig:NMF_parts}
\end{figure}

$L_{s}$ is backpropagated through the PAM to the PartNet and AFD for improving the predictions $P$ and $\Pi$ such that $L_{s}$ is minimized. We regularize this learning by using the NMF of the ground-truth instance segmentation masks as follows.  All ground-truth segmentations of all object classes in the training dataset $\{m_n^*:n=1,\dots,D\}$ are first re-sized and stacked into a large $(H_r W_r) \times D$ matrix $M^*$. Then, we apply the NMF as $M^* \approx P^*U$, where $P^*$ is the non-negative basis matrix with size $(H_r W_r) \times J$, and $U$ is the non-negative weight matrix with size $J \times D$. The product $P^*U$ is a low-rank approximation of $M^*$.
Due to the non-negative constraint, $P^*$ is forced to be sparse. Fig.~\ref{fig:NMF_parts} shows the NMF's bases in $P^*$ that we computed on the ground-truth masks $\{m_n^*\}$ of the COCO-$20^i$ dataset \cite{michaelis2018one}. Hence, $P^*$ is interpreted as storing parts  of all training object classes in $M^*$. Conveniently, the NMF, i.e., $P^*$, can be pre-computed before learning.

We use $P^*$ to regularize our learning of the PartNet such that it produces part maps $P^+_n$ that are similar to $P^*$. Note that the $J$ latent parts in $P^+_n$ may be differently indexed from the NMF's $J$ bases stored in $P^*$. Therefore, we employ the Hungarian algorithm \cite{kuhn1955hungarian} to identify the one-to-one correspondence between the latent parts $P_{n,j}^+$ and the bases $P_{j'}^*$, $j'=\text{Hungarian}(j)$, $j=1,\dots,J$, based on their 
intersection-over-union (IoU) score. With this correspondence, we specify the NMF regularization loss as%
\begin{equation}
\label{eq:PartLoss}
    L_{\text{NMF}} = \frac{1}{N}\sum_{n=1}^N \sum_{j=1}^J l_{dice}(P_{n,j}^+, P^*_{j'}) ,\quad j'=\text{Hungarian}(j).
\end{equation} 

The total loss for training our \Approach~is specified as
\begin{equation}
\label{eq:total_loss}
    L = \lambda_1 L_{c} + \lambda_2 L_{b} + \lambda_3 L_{s} + \lambda_4 L_{\text{NMF}}
\end{equation} 
where the $\lambda$'s are positive coefficients, specified in Sec.~\ref{sec:implement}.

\section{Experiments}
\label{sec:experiments}

\textbf{Dataset \& Metrics}: For evaluation, we conduct the standard 4-fold cross-validation on the benchmark COCO-$20^i$ dataset \cite{michaelis2018one}. In each fold $i=0,...,3$, we sample 20  test classes from the 80 object classes in MSCOCO \cite{lin2014microsoft}, and use the remaining 60 classes for training. For each COCO-$20^i$, we randomly sample five separate test sets, and report the average results as well as their standard deviations, as in   \cite{michaelis2018one}.  We follow the standard testing protocol for 1-way $K$-shot instance segmentation  \cite{michaelis2018one}. Our training uses $K=1$ support image, while in testing we consider two settings with $K=1$ and $K=5$  support images. For each test query, $K$ support images are  randomly sampled from the remaining test images showing the same test class as the query. 
We use the evaluation tool provided by the COCO dataset to compute the following metrics:  mAP50 of both segmentation and detection is computed at a single intersection-over-union (IoU) threshold of $50\%$ between our prediction and ground truth; mAR10 is average recall given 10 detections per image.
Our results of other metrics for each fold can be found in the supplementary material.

\subsection{Implementation Details}
\label{sec:implement}

Our backbone CNN is ResNet-50 \cite{he2015deep} and the FPN of \cite{lin2017feature}, as in prior work.  \Approach~is implemented using the mmdetection toolbox \cite{chen2019mmdetection}. We empirically find that using the P3 to P5 feature map levels gives the best results. 
The FG scoring and part-importance prediction share the first two convolutional layers with GroupNorm \cite{wu2018group} and ReLU activation in between, but use two separate heads for binary class and part-importance predictions. We strictly follow the design of classification head and box regression head of FCOS \cite{tian2019fcos}. SimNet has a block of four fully connected layers with BatchNorm \cite{ioffe2015batch} and ReLU in between each layer, followed by a top convolutional layer, where the block predicts the weights of the top layer.
The PartNet consists of 5 convolutional layers with GroupNorm and ReLU, and an Upsampling layer to upscale the resolution by two in the middle of layer 3 and 4, following the design of mask head of Mask-RCNN \cite{he2017mask}. 
For learning, we use SGD with momentum \cite{cortes1995support} with the learning rate of $1e^{-3}$.
The number of training epochs is 12, which is similar to the setting of 1x in Mask-RCNN. The mini-batch size is 16. The query images are resized to $800{\times}1333$ pixels. The support images and masks are cropped around the ground-truth bounding boxes and re-sized to $128 \times 128$ pixels. We set $H_{\text{r}}=W_{\text{r}}=32$,  $\alpha =0.25$ and $\gamma =2$ in \eqref{eq:cls_loss_function}, and  $\lambda_1 = \lambda_2 = \lambda_3 = 1, \lambda_4 = 0.1$ in \eqref{eq:total_loss}.

\subsection{Ablation Study}
Ablations and sensitivity to input parameters are evaluated for the setting $K=1$.

{\bf Analysis of the number of parts.}
Tab.~\ref{tab:num_parts} reports how the number of latent parts $J$ affects our results. We vary $J$ while keeping other hyper-parameters unchanged. From Tab.~\ref{tab:num_parts}, $J=16$ gives the best performance. When $J$ is small, \Approach~cannot reliably capture object variations, and when $J$ approaches the number of classes considered in training, the PartNet tends to predict shape prototypes as in YOLACT \cite{bolya2019yolact}, instead of parts. Therefore, we use $J=16$ for all other evaluations. 

{\bf  Analysis of the predicted part importance.} 
Tab.~\ref{tab:sparsity} shows a percentage of the latent parts whose predicted importance for segmentation is higher than a threshold $\theta \in (0, 1)$. As can be seen, for a given object instance, most latent parts are usually estimated as irrelevant. That is, the PAM essentially uses only a few most important latent parts to form the instance segmentation mask.  

\begin{table}[]
\begin{center}
\small
\begin{tabular}{c|c|c|c|c|c|c|c}
\hline
\hline
\textbf{\# parts}  & 1 & 2 & 4 & 8 & 16 & 32 & 64  \\
\hline
mAP50 & 16.3 & 17.2 & 17.9 & 18.4 & \textbf{18.8} & 18.5 & 18.0 \\
\hline 
\hline
\end{tabular}
\end{center}
\caption{mAP50 for segmentation of \Approach~on COCO-$20^0$ for different numbers of the latent parts $J$ used.
}
\label{tab:num_parts}
\end{table}

\begin{table}[]
\begin{center}
\small
\begin{tabular}{c|c|c|c|c|c}
\hline
\hline
$\theta$ & 0.1 & 0.3 & 0.5 & 0.7 & 0.9 \\
\hline
\% parts with $\sigma(\pi) {>}\theta$ & 0.32 & 0.29 & 0.26 & 0.21 & 0.09  \\
\hline 
\hline
\end{tabular}
\end{center}
\caption{The predicted importance of most latent parts is on average lower than a threshold $\theta\in(0,1)$, so instance segmentation is formed from only a few important parts.}
\label{tab:sparsity}
\end{table}

\begin{table}[]
\begin{center}
\small
\begin{tabular}{c|c|c|c|c}
\hline
\hline
\multirow{2}{*}{\textbf{Ablations}} & \multicolumn{2}{c|}{\textbf{Object detection}} & \multicolumn{2}{c}{\textbf{Instance segm}} \\ \cline{2-5} 
 & \textbf{mAP50} & \textbf{mAR10} & \textbf{mAP50} & \textbf{mAR10}   \\ \hline
FIS         & 18.4 & 25.0 & 16.4 & 21.3 \\
FAIS        & \textbf{20.9} & \textbf{28.1} & 18.0 & 22.5  \\
FAIS-SimNet & 20.2 & 26.9 & 17.4 & 22.0 \\
FAIS-CWM    & 16.5 & 24.3 & 15.4 & 20.5 \\
FPIS        & 18.4 & 25.0 & 17.4 & 22.4 \\
FAPIS-$L_{\text{NMF}}$& 20.5 & 27.5 & 18.4 & 22.7 \\
FAPIS       & \textbf{20.9} & \textbf{28.1} & \textbf{18.8} & \textbf{23.3}  \\
\hline 
\hline
\end{tabular}
\end{center}
\caption{mAP50 and mAR10 of one-shot object detection and instance segmentation on COCO-$20^0$ for different variants of FAPIS. The  PIS  is used only for instance segmentation, so the variants FIS and FPIS have lower mAP50 for object detection. The AFD is used for both instance detection and segmentation, and thus FAIS and FAPIS have higher mAP50 for both object detection and instance segmentation than FIS and FPIS.
The AFD without SimNet decreases its discriminative power so FAIS-SimNet has lower mAP50 for object detection than FAIS.
When the channel-wise multiplication (CWM) is replaced with SimNet alone, performance of FAIS significantly decreases, and becomes even lower than that of FIS. This shows that SimNet does not replace but extends CWM.
FAPIS-$L_{\text{NMF}}$ has lower mAP50 for instance segmentation than FAPIS. 
}
\label{tab:variants}
\end{table}

{\bf Ablations of \Approach.}
Tab.~\ref{tab:variants} demonstrates how the AFD and PIS affect performance of the following six ablations: 
\begin{itemize}[itemsep=-2pt,topsep=-2pt, partopsep=-1pt]
    \item FIS: AFD is replaced with the region proposal network (RPN) \cite{ren2015faster} for  anchor-based detection; and PIS is replaced with the standard mask-head of Mask-RCNN.
    \item FAIS: PIS replaced with mask-head of Mask-RCNN. 
    \item FAIS-SimNet: FAIS without SimNet, and FG scores are predicted as in \cite{tian2019fcos} (see Fig.~\ref{fig:feature_extraction}).
    \item FAIS-CWM: FAIS without channel-wise multiplication (CWM).
    \item FPIS: AFD is replaced with RPN \cite{ren2015faster}
    \item FAPIS-$L_{\text{NMF}}$: FAPIS trained without the regularization loss  $L_{\text{NMF}}$.
    \item \Approach: our approach depicted in Fig.~\ref{fig:part_assembling_module}
\end{itemize}
The top three variants test our contributions related to anchor-free detection, and the last three test our contributions related to modeling the latent parts.

\begin{table*}[]

\begin{center}
\small
\begin{tabular}{c|c|c|c|c|c|c}
\hline
\hline
\textbf{$\#$ shots} & \textbf{Method}  & \textbf{COCO-$20^0$}      & \textbf{COCO-$20^1$}      & \textbf{COCO-$20^2$ }     & \textbf{COCO-$20^3$}      & mean  \\
\hline
\multirow{4}{*}{K=1} & Meta-RCNN \cite{yanICCV19metarcnn}        & 17.7 $\pm$ 0.7 & 19.2 $\pm$ 0.6 & 17.7 $\pm$ 0.3 & 21.1 $\pm$ 0.4 & 18.9 \\
& Siamese M-RCNN \cite{michaelis2018one} & 18.3 $\pm$ 0.8 & 19.5 $\pm$ 0.7 & 18.0 $\pm$ 0.4 & 21.5 $\pm$ 0.6 & 19.3 \\
& YOLACT \cite{bolya2019yolact}  &    18.0 $\pm$ 0.5 &  18.8 $\pm$ 0.5 & 17.8 $\pm$ 0.6 & 21.2 $\pm$ 0.7 & 19.0 \\
& \Approach~ & \textbf{20.9  $\pm$ 0.4}  &  \textbf{20.4 $\pm$ 0.1} & \textbf{20.0 $\pm$ 0.6} & \textbf{23.4 $\pm$ 0.5}   &  \textbf{21.2}   \\ 

\hline

\multirow{4}{*}{K=5} & Meta-RCNN \cite{yanICCV19metarcnn}        & 19.1 $\pm$ 0.4 & 21.2 $\pm$ 0.2 & 19.6 $\pm$ 0.5 & 24.0 $\pm$ 0.2 & 21.0 \\
& Siamese M-RCNN \cite{michaelis2018one} & 20.0 $\pm$ 0.4 & 21.6 $\pm$ 0.3 & 20.2 $\pm$ 0.4 & 24.1 $\pm$ 0.3 & 21.5 \\
& YOLACT \cite{bolya2019yolact} &  20.8 $\pm$ 0.4 & 21.1 $\pm$ 0.2 & 20.2 $\pm$ 0.5 & 24.8 $\pm$ 0.2 & 21.7 \\
& \Approach~   & \textbf{22.6 $\pm$ 0.3}  &  \textbf{22.8 $\pm$ 0.0} & \textbf{22.6 $\pm$ 0.6} & \textbf{26.4 $\pm$ 0.2} & \textbf{23.6} \\

\hline 
\hline
\end{tabular}
\end{center}
\vspace{-5pt}
\caption{mAP50 with standard deviation of one-shot and five-shot {\bf object detection} on COCO-$20^i$. The best results are in bold.}
\label{tab:few_shot_det}
\end{table*}

\begin{table*}[]
\begin{center}
\small
\begin{tabular}{c|c|c|c|c|c|c}
\hline
\hline
\textbf{$\#$ shots} & \textbf{Method} & 
\textbf{COCO-$20^0$}      & \textbf{COCO-$20^1$}      & \textbf{COCO-$20^2$ }     & \textbf{COCO-$20^3$}      & mean  \\
\hline

\multirow{4}{*}{K=1} & Meta-RCNN \cite{yanICCV19metarcnn}        & 16.0 $\pm$ 0.6 & 16.1 $\pm$ 0.5 & 15.8 $\pm$ 0.3 & 18.6 $\pm$ 0.4 & 16.6 \\
& Siamese M-RCNN \cite{michaelis2018one} & 16.6 $\pm$ 0.8 & 16.6 $\pm$ 0.6 & 16.3 $\pm$ 0.7 & 19.3 $\pm$ 0.6 & 17.2 \\
& YOLACT \cite{bolya2019yolact}  &    16.8 $\pm$ 0.6 &  16.5 $\pm$ 0.5 & 16.1 $\pm$ 0.4 & 19.0 $\pm$ 0.6 & 17.1 \\
& \Approach~  &  \textbf{18.8 $\pm$ 0.3} &  \textbf{17.7 $\pm$ 0.1} & \textbf{18.2 $\pm$ 0.7} & \textbf{21.4 $\pm$ 0.4} & \textbf{19.0} \\

\hline 

\multirow{4}{*}{K=5} & Meta-RCNN \cite{yanICCV19metarcnn}        & 17.4 $\pm$ 0.3 & 17.8 $\pm$ 0.2 & 17.7 $\pm$ 0.7 & 21.3 $\pm$ 0.2 & 18.6 \\
& Siamese M-RCNN \cite{michaelis2018one} & 17.5 $\pm$ 0.4 & 18.5 $\pm$ 0.1 & 18.2 $\pm$ 1.0 & 22.4 $\pm$ 0.2 & 19.2 \\
& YOLACT \cite{bolya2019yolact} &  17.6 $\pm$ 0.2 & 18.4 $\pm$ 0.2 & 17.9 $\pm$ 0.6 & 21.8 $\pm$ 0.3 & 18.9 \\
& \Approach~ & \textbf{20.2 $\pm$ 0.2} & \textbf{20.0 $\pm$ 0.1} & \textbf{20.4 $\pm$ 0.7} & \textbf{24.3 $\pm$ 0.2} & \textbf{21.2}\\

\hline 
\hline
\end{tabular}
\end{center}
\vspace{-5pt}
\caption{mAP50 with standard deviation of one-shot and five-shot {\bf instance segmentation} on COCO-$20^i$. The best results are in bold.}
\label{tab:few_shot_seg}
\end{table*}

From Tab.~\ref{tab:variants}, our AFD in FAIS improves performance by $1.6\%$ on mAP50 over FIS that uses the anchor-based detector.  
However, removing the SimNet from the AFD in FAIS-SimNet decreases performance, justifying our SimNet as a better way to predict FG scores than in \cite{yanICCV19metarcnn,michaelis2018one}.
The PIS in FPIS gives a performance gain of $1\%$ on mAP50 in instance segmentation over FIS. Our performance decreases when \Approach~is trained without the NMF regularization in FAPIS-$L_{\text{NMF}}$. \Approach~gives the best performance in comparison with the strong baselines.

Note that we cannot directly evaluate our part detection, since we do not have ground-truth annotations of parts.


\begin{figure}[h!]
    \centering
    \includegraphics[scale=0.235]{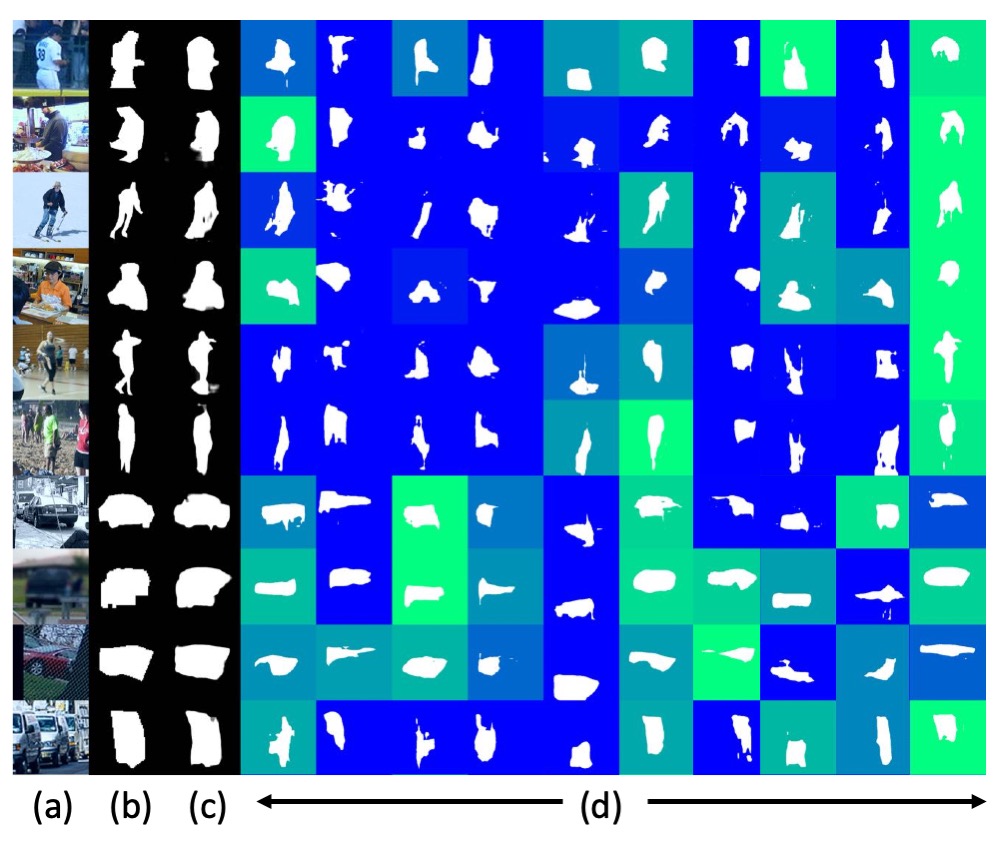}
    \caption{Visualization of the most important 10 latent parts out of 16 predicted for example instances of the ``human'' and ``car'' classes from the COCO-$20^0$ validation dataset. From left to right: (a) input image, (b) GT segmentation, (c) predicted segmentation, (d) 10 most relevant parts. The predicted importance of the parts is color-coded from blue (smallest) to green (largest).}
    \label{fig:qualitative_parts_16}
\end{figure}

\begin{figure*}[h!]
    \centering
    \includegraphics[scale=0.37]{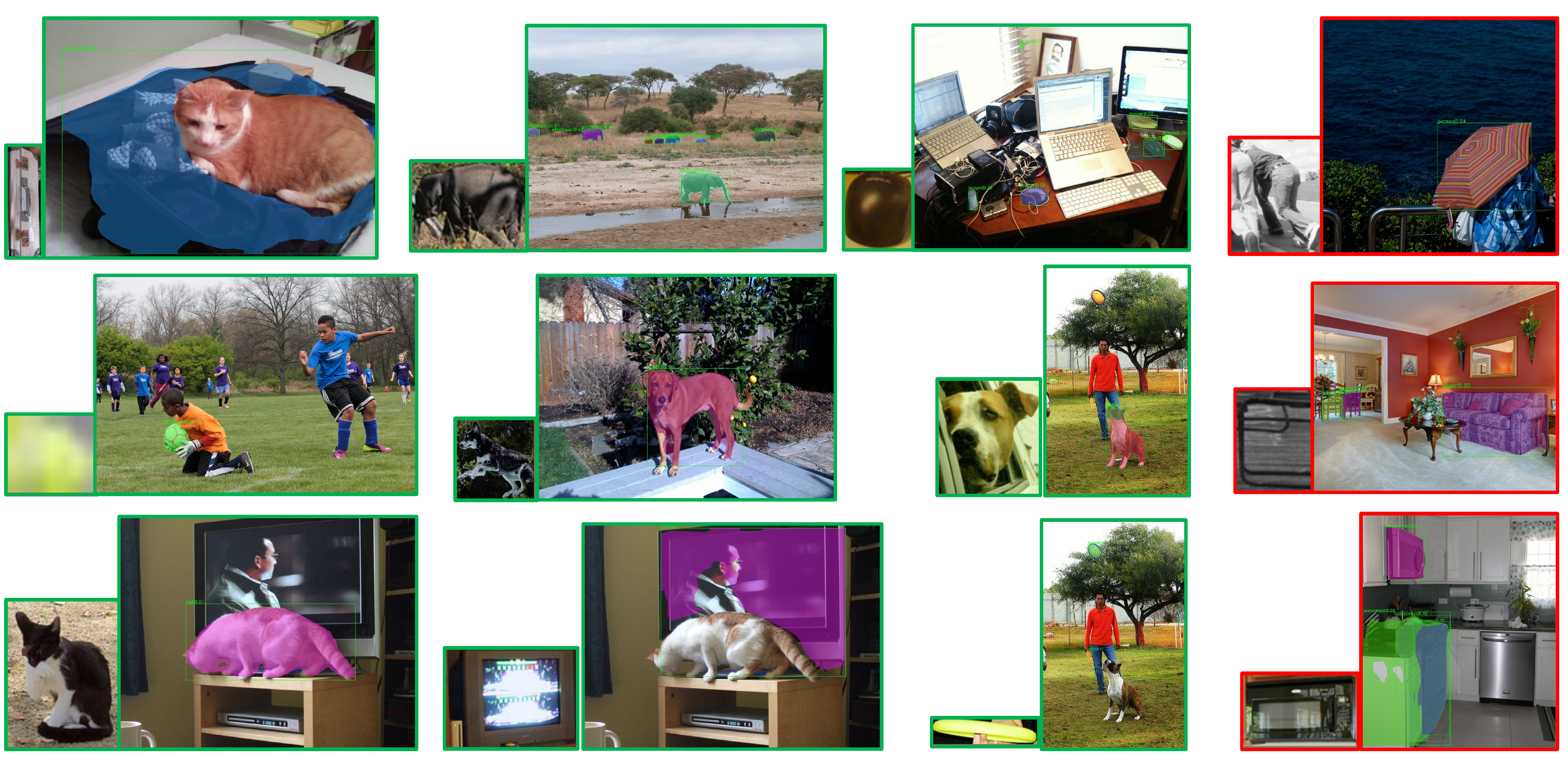}
    \caption{Our one-shot instance segmentation on COCO-$20^0$. For each pair of images, the support is the smaller and the query is the larger image. Results for the segmentation and bounding-box detection of each instance are marked with  distinct colors in the query. The green border indicates success, and the red border marks failure. \Approach~typically fails when instances in the support image are very different in appearance, shape, or 3D pose from instances in the query.  Best viewed in color.}
    \label{fig:qualitative_part0}
\end{figure*}

\subsection{Comparison with prior work}
\Approach~is compared with Meta-RCNN \cite{yanICCV19metarcnn}, Siamese Mask-RCNN \cite{michaelis2018one} and YOLACT \cite{bolya2019yolact}. Meta-RCNN and Siamese Mask-RCNN use the same anchor-based framework, but differ in that Meta-RCNN performs the feature correlation after the RPN, whereas Siamese Mask-RCNN does this before the RPN. YOLACT is another anchor-based instance segmentor based on RetinaNet \cite{lin2017focal}, which learns to predict object shapes instead of object parts, and we adapt it to few-shot instance segmentation by changing its feature extraction module as in Fig.~\ref{fig:feature_extraction}.

Note that the results of Siamese Mask-RCNN  in \cite{michaelis2018one} are reported for evaluation where the support images do not provide ground-truth segmentation masks, but only bounding boxes. Therefore, for fairness, we evaluate Siamese Mask-RCNN (without any changes to their public code) on the same input as for Meta-RCNN, YOLACT, and \Approach -- i.e., when the support images provide ground-truth segmentation masks of object instances. In addition, in \cite{yanICCV19metarcnn}, the results of Meta-RCNN are reported for a different version of COCO. Therefore, we evaluate Meta-RCNN (without any changes to their public code) on our test sets.

Tab.~\ref{tab:few_shot_det} reports one-shot and five-shot object detection results, and Tab.~\ref{tab:few_shot_seg} shows one-shot and five-shot instance segmentation results. From these tables, Siamese Mask-RCNN performs slightly better than Meta-RCNN, since the feature correlation happens before the RPN, enabling the RPN to better adapt its bounding box proposals to the target class. 
Our \Approach~outperforms YOLACT by $2\%$ for one-shot and five-shot instance segmentation. This demonstrates  advantages of using latent parts and their layouts for assembling object shapes in our approach over the YOLACT's prototypes in few-shot instance segmentation. 

Our gains in performance are significant in the context of previous performance improvements reported by prior work, where gains were only by about 0.6\% by \cite{michaelis2018one} vs \cite{yanICCV19metarcnn}, by 0.1\% by \cite{michaelis2018one} vs \cite{bolya2019yolact}, by 0.5 \% \cite{bolya2019yolact} vs \cite{yanICCV19metarcnn}. We improve by 1.8\% and 2.0\% over \cite{michaelis2018one} on one-shot and five-shot instance segmentation, respectively.


\subsection{Qualitative Evaluation}
Fig.~\ref{fig:qualitative_parts_16} visualizes the most important 10 of 16 latent parts and their relative importance for a few example instances of the ``human'' and ``car'' classes from the COCO-$20^0$ validation set. The figure shows that the predicted latent parts are smaller components of the objects, and that some parts may be assigned a meaningful interpretation.  Our visualization of the latent parts for other classes can be found in the supplementary material.

A few representative success and failure results on COCO-$20^0$ are illustrated in Fig.~\ref{fig:qualitative_part0}, and others are included in the supplementary material. 

\section{Conclusion}
\label{sec:conclusions}
We have specified a new few-shot anchor-free part-based instance segmenter (\Approach) that predicts latent object parts for instance segmentation, where part annotations are not available in training. \Approach~uses our new anchor-free object detector (AFD), PartNet for localizing the latent parts, and part assembly module for fusing the part activation maps by their predicted importance into a segmentation of every detected instance. We have regularized learning such that the identified latent parts and similar to the sparse bases of the non-negative matrix factorization of the ground-truth segmentation masks. Our evaluation on the benchmark COCO-$20^i$ dataset  demonstrates that: we significantly outperform the state of the art; our prediction of latent parts gives better performance than using the YOLACT's prototypes or using the standard mask-head of Mask-RCNN; and our anchor-free AFD improves performance over the common anchor-based bounding-box prediction. 

\vspace{10pt}

\noindent{\bf{Acknowledgement}}. This work was supported in part by DARPA XAI Award N66001-17-2-4029 and DARPA MCS Award N66001-19-2-4035.

{\small
\bibliographystyle{ieee_fullname}
\bibliography{egbib}
}

\onecolumn

\begin{center}
   \Large \textbf{Supplementary Material}
\end{center}


\textbf{1. Full COCO metrics of \Approach:}
Tab.~\ref{tab:coco_bbox} and Tab.~\ref{tab:coco_segm} show the average over five runs of full COCO metrics of \Approach~on object detection and instance segmentation respectively.

\begin{table*}[h]
\begin{center}
\small
\setlength{\tabcolsep}{5pt}
\begin{tabular}{ r  |l|ccc|ccc|ccc|ccc}
\hline 
\hline
\textbf{$\#$ shots} & \textbf{Part} & \textbf{AP} & \textbf{AP$_{50}$} & \textbf{AP$_{75}$} & \textbf{AP$_S$} & \textbf{AP$_M$} & \textbf{AP$_L$} & \textbf{AR$_{1}$} & \textbf{AR$_{10}$} & \textbf{AR$_{100}$} & \textbf{AR$_{S}$} & \textbf{AR$_{M}$} & \textbf{AR$_{L}$} \\
 \hline
\multirow{4}{*}{K=1}  
&  COCO-$20^0$     & 12.8 & 20.9 & 13.2 & 8.7 & 13.9 & 18.8 & 11.9 & 28.1 & 38.8 & 24.0 & 41.7 & 51.9 \\
&  COCO-$20^1$     & 12.0 & 20.4 & 12.3 & 6.9 & 14.1 & 17.2 & 12.3 & 31.5 & 44.6 & 31.2 & 51.3 & 59.7 \\
&  COCO-$20^2$     & 11.6 & 20.0 & 11.7 & 9.0 & 10.7 & 19.1 & 10.9 & 28.8 & 42.9 & 31.7 & 46.6 & 57.6 \\
&  COCO-$20^3$     & 15.1 & 23.4 & 15.9 & 10.5 & 13.7 & 19.1 & 14.4 & 33.7 & 46.1 & 32.4 & 49.8 & 59.5 \\
\hline
\multirow{4}{*}{K=5}  
&  COCO-$20^0$     & 14.1 & 22.6 & 14.6 & 8.3 & 14.8 & 22.0 & 13.2 & 28.9 & 39.4 & 24.5 & 41.9 & 54.0 \\
&  COCO-$20^1$     & 13.4 & 22.8 & 13.7 & 7.1 & 15.9 & 19.6 & 13.8 & 33.1 & 45.8 & 33.1 & 53.8 & 61.0 \\
&  COCO-$20^2$     & 13.0 & 22.6 & 13.4 & 9.6 & 13.0 & 21.2 & 12.0 & 31.0 & 43.9 & 33.3 & 47.6 & 59.1 \\
&  COCO-$20^3$     & 16.9 & 26.4 & 18.0 & 11.3 & 15.3 & 23.8 & 15.2 & 35.0 & 47.3 & 33.1 & 51.4 & 61.7 \\
\hline
\hline
\end{tabular}
\end{center}
\vspace{-5pt}
\caption{Object detection results of \Approach~on full COCO metrics}
\label{tab:coco_bbox}
\end{table*}
\vspace{-10pt}
\begin{table*}[h]
\begin{center}
\small
\setlength{\tabcolsep}{5pt}
\begin{tabular}{ r  |l|ccc|ccc|ccc|ccc}
\hline
\hline
\textbf{$\#$ shots} & \textbf{Part} & \textbf{AP} & \textbf{AP$_{50}$} & \textbf{AP$_{75}$} & \textbf{AP$_S$} & \textbf{AP$_M$} & \textbf{AP$_L$} & \textbf{AR$_{1}$} & \textbf{AR$_{10}$} & \textbf{AR$_{100}$} & \textbf{AR$_{S}$} & \textbf{AR$_{M}$} & \textbf{AR$_{L}$} \\
 \hline
\multirow{4}{*}{K=1}  
&  COCO-$20^0$     & 10.2 & 18.8 & 10.1 & 5.7 & 9.7 & 18.0 & 9.8 & 23.3 & 31.4 & 18.9 & 33.6 & 42.8 \\
&  COCO-$20^1$     & 8.9 & 17.7 & 8.0 & 3.8 & 10.4 & 15.7 & 10.1 & 26.0 & 36.7 & 23.0 & 44.2 & 50.6 \\
&  COCO-$20^2$     & 9.8 & 18.2 & 9.7 & 4.3 & 8.9 & 18.3 & 10.5 & 24.8 & 35.9 & 26.2 & 39.6 & 48.7 \\
&  COCO-$20^3$     & 12.5 & 21.4 & 13.3 & 7.1 & 11.4 & 19.2 & 12.6 & 28.3 & 37.4 & 22.9 & 41.4 & 52.8\\
\hline
\multirow{4}{*}{K=5}  
&  COCO-$20^0$     & 11.3 & 20.2 & 10.9 & 5.0 & 10.5 & 20.8 & 10.9 & 24.0 & 32.1 & 18.9 & 33.9 & 45.8 \\
&  COCO-$20^1$     & 10.3 & 20.0 &  9.4 & 3.8 & 11.9 & 19.0 & 11.2 & 27.6 & 38.4 & 24.4 & 46.0 & 54.7 \\
&  COCO-$20^2$     & 11.4 & 20.4 & 12.1 & 6.5 & 10.6 & 20.4 & 11.3 & 25.5 & 35.5 & 24.4 & 39.3 & 47.9 \\
&  COCO-$20^3$     & 14.3 & 24.3 & 15.0 & 7.5 & 13.3 & 22.9 & 13.5 & 30.1 & 39.6 & 24.8 & 43.8 & 54.3\\
\hline
\hline
\end{tabular}
\end{center}
\vspace{-5pt}
\caption{Instance segmentation results of \Approach~on full COCO metrics}
\label{tab:coco_segm}
\end{table*}

\begin{figure*}[h!]
    \centering
    \includegraphics[scale=0.4]{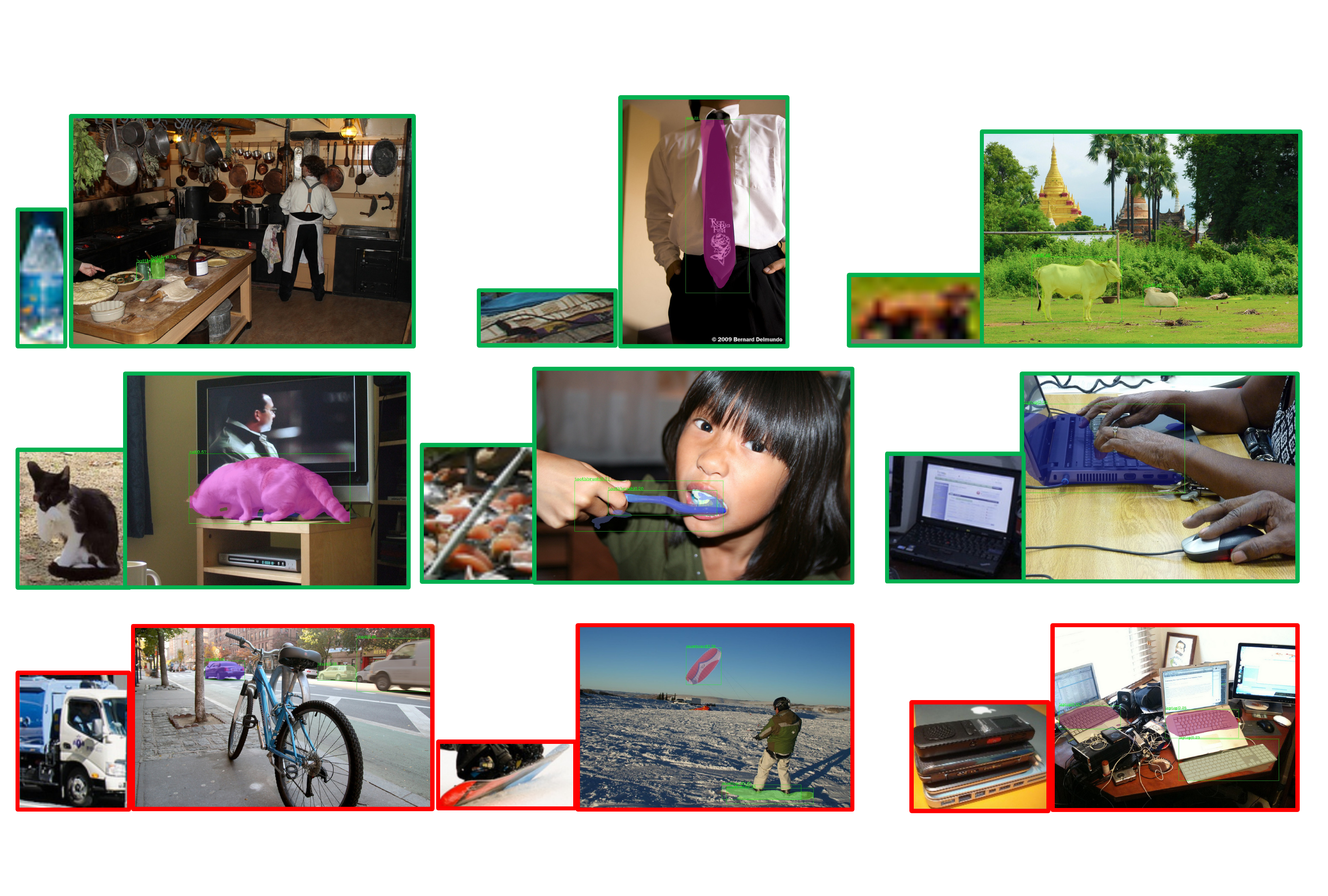}
    \caption{Our one-shot instance segmentation on COCO-$20^1$.}
    \label{fig:qualitative_part1}
\end{figure*}

\begin{figure*}[h!]
    \centering
    \includegraphics[scale=0.4]{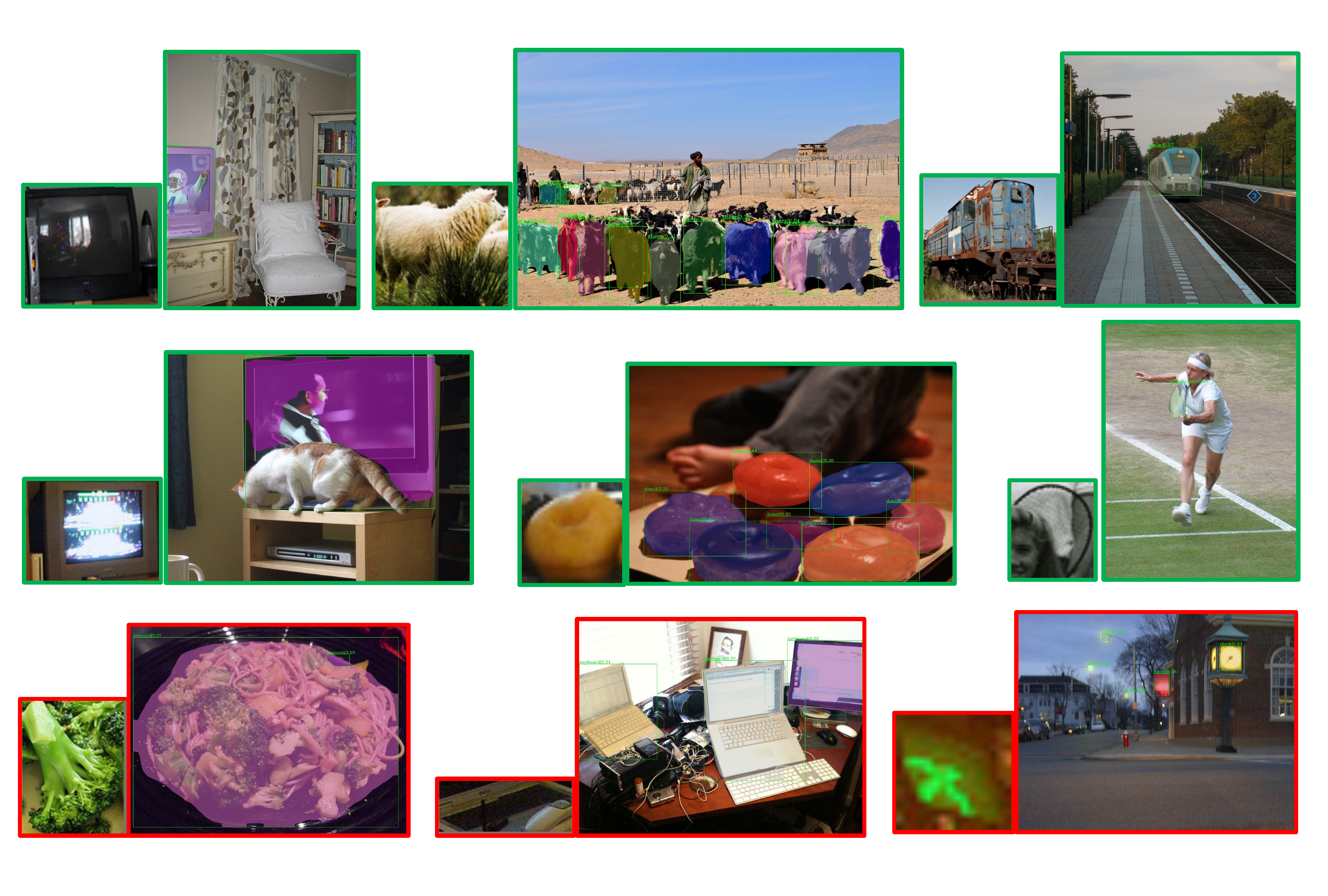}
    \caption{Our one-shot instance segmentation on COCO-$20^2$.}
    \label{fig:qualitative_part2}
\end{figure*}

\begin{figure*}[h!]
    \centering
    \includegraphics[scale=0.4]{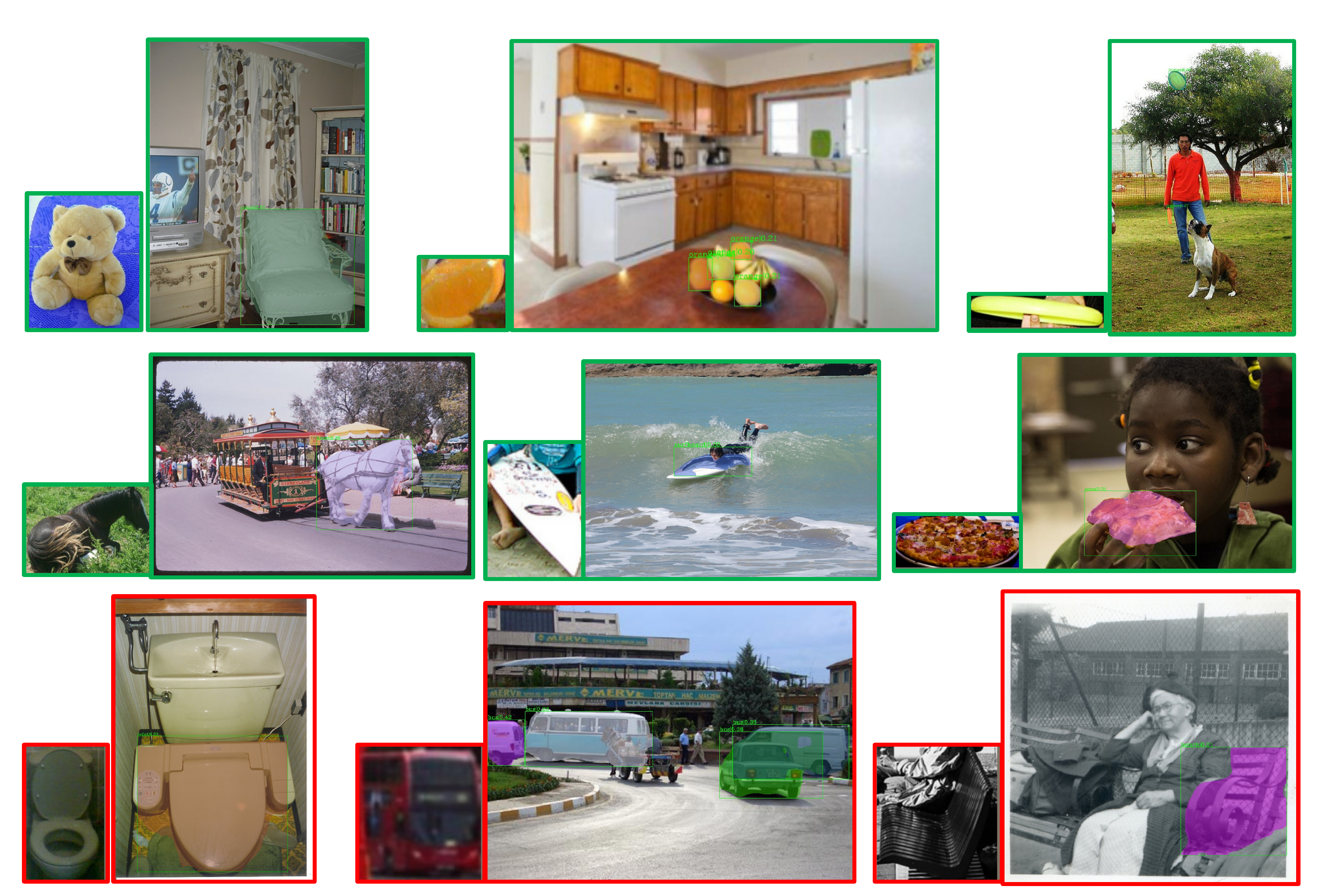}
    \vspace{10pt}
    \caption{Our one-shot instance segmentation on COCO-$20^3$.}
    \label{fig:qualitative_part3}
\end{figure*}

\textbf{2. Further qualitative results: }
We show more qualitative results on COCO-$20^i, i=1..3$ in 
Fig.~\ref{fig:qualitative_part1},
Fig.~\ref{fig:qualitative_part2},
Fig.~\ref{fig:qualitative_part3}, respectively. 
For each pair of images, the small one is the support and the large one is the query. Each instance segmentation result is marked with a different color in the query. The green border indicates success cases, and the red border marks failure cases. FAPIS typically fails when instances in the support image are very different in appearance, shape, or 3D pose from instances in the query.  Best viewed in color.


\textbf{3. Visualization of 16 latent parts:}
We show visualization of learned parts and their weights contributing to the final masks from COCO-$20^0$ validation set in Fig.~\ref{fig:qualitative_parts_16}.

\begin{figure*}[h!]
    \centering
    \includegraphics[scale=0.8]{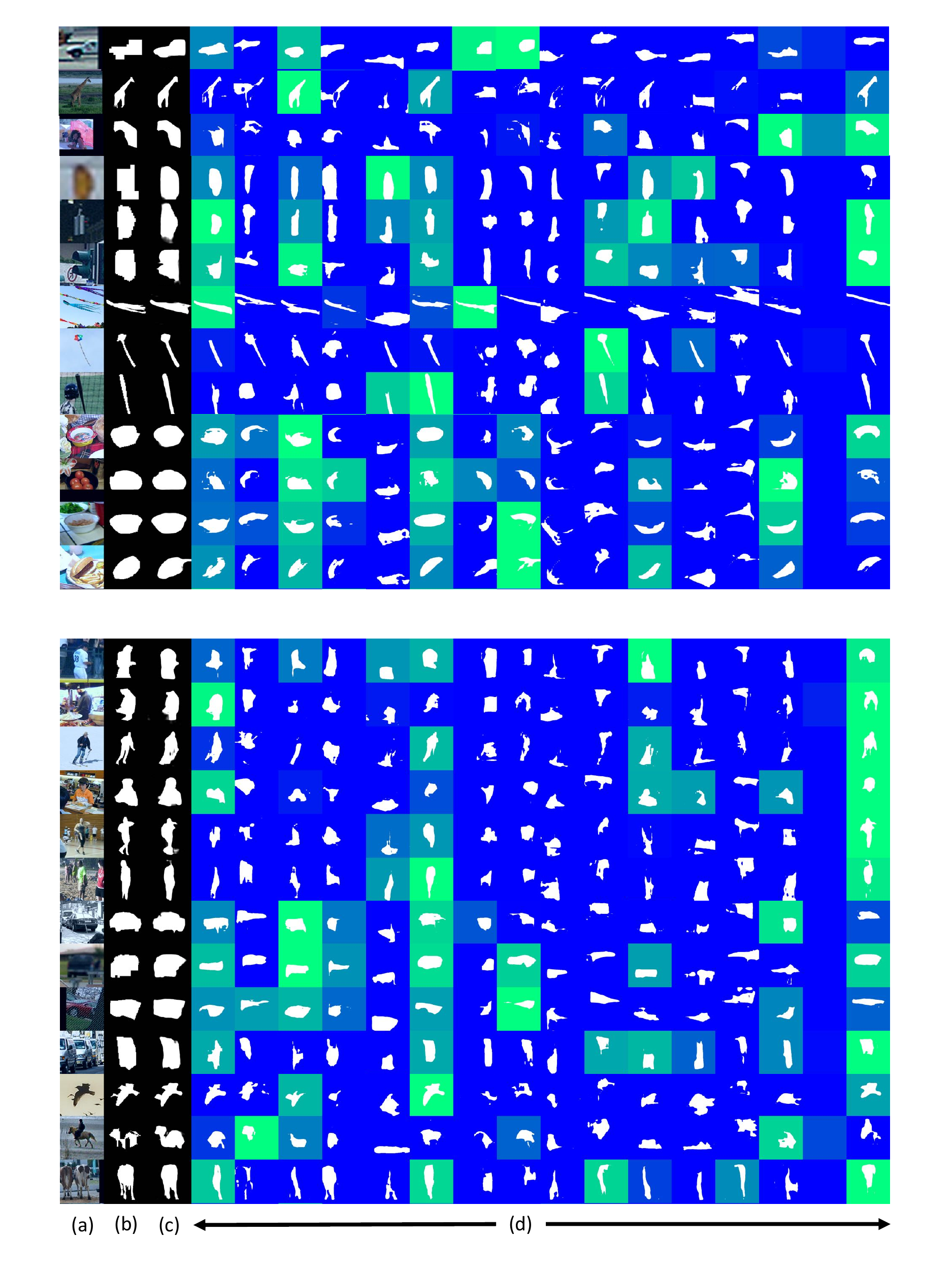}
    \vspace{-10pt}
    \caption{Visualization of 16 latent parts for both training (upper) and test (lower) classes from COCO-$20^0$ validation set. From left to right: (a) input image, (b) GT segmentation, (c) predicted segmentation, (d) 16 predicted object parts. The predicted importance of the parts is color-coded from blue (smallest) to green (largest).}
    \label{fig:qualitative_parts_16}
\end{figure*}

\end{document}